\documentclass[a4paper]{article}
\usepackage{booktabs, multirow} 
\usepackage{soul}
\usepackage[table]{xcolor} 
\usepackage{changepage,threeparttable} 
\usepackage{float}
\usepackage{pgfplots, pgfplotstable}
\usepgfplotslibrary{colorbrewer}
\pgfplotsset{compat=1.15}
\usepackage[hyphens]{url}
\usepackage{INTERSPEECH2021}

\newcommand\blfootnote[1]{%
  \begingroup
  \renewcommand\thefootnote{}\footnote{#1}%
  \addtocounter{footnote}{-1}%
  \endgroup
}

\title{Improving Low-Resource Speech Recognition with Pretrained Speech Models: Continued Pretraining vs. Semi-Supervised Training}
\name{Mitchell DeHaven, Jayadev Billa}
\address{
  Information Sciences Institute - University of Southern California}
\email{mdehaven@isi.edu, jbilla@isi.edu}

\begin{document}

\maketitle
\begin{abstract}
Self-supervised Transformer based models, such as wav2vec 2.0 and HuBERT, have produced significant improvements over existing approaches to automatic speech recognition (ASR). This is evident in the performance of the wav2vec 2.0 based pretrained XLSR-53 model across many languages when fine-tuned with available labeled data. However, the performance from finetuning these models can be dependent on the amount of in-language or similar-to-in-language data included in the pretraining dataset. In this paper we investigate continued pretraining (CoPT) with unlabeled in-language audio data on the XLSR-53 pretrained model in several low-resource languages. CoPT is more computationally efficient than semi-supervised training (SST), the standard approach of utilizing unlabeled data in ASR, since it omits the need for pseudo-labeling of the unlabeled data. We show CoPT results in word error rates (WERs), equal to or slightly better than using SST. In addition, we show that using the CoPT model for pseudo-labeling, and using these labels in SST, results in further improvements in WER. 
\end{abstract}
\noindent\textbf{Index Terms}: Continued Pretraining (CoPT), Transformer, self-supervision, speech recognition

\blfootnote{This work was submitted for publication to Interspeech 2022.}

\section{Introduction}

The success of self-supervision trained Transformer models such as GPT~\cite{radford2018improving,radford2019language}, BERT~\cite{DBLP:journals/corr/abs-1810-04805} and others in natural language processing (NLP) has led to their adoption in computer vision (e.g. \cite{vit}) and audio processing (e.g. \cite{wav2vec}). The ability to use large pools of unlabeled data in Transformer model training appears to be a key driver of performance on a variety of downstream tasks, especially when operating with little supervised training data.

In both machine translation and speech recognition, large pretrained Transformers, trained on multilingual data, learn representations that are broadly useful across many languages, as opposed to models trained on only English data~\cite{xlm, xlsr-53}. These models have been able to improve machine translation and speech recognition on low-resource languages that otherwise have very poor performance. However, for many of these pretrained multilingual models a few languages dominate the pretraining data set, particularly English.

Intuitively, the amount of in-language data in the pretraining dataset should impact the models' subsequent performance on the language in downstream tasks. Indeed, Wu and Dredze~\cite{wu-dredze} show that the most under-represented languages in a multilingual BERT model perform significantly worse on downstream tasks than better-represented languages in the pretraining set. Thus, an approach to improving performance in low-resource languages is to increase its representation during pretraining. However, given the size of the datasets and models, it is computationally expensive to pretrain these models from scratch with more of, or exclusively on, target language data in the pretraining dataset, although it has been explored extensively in other domains~\cite{biobert, bertweet, scibert}. In this work, we employ an alternative to this for speech recognition: simply continue the pretraining task with unsupervised audio from the target language.

The idea of continued pretraining (CoPT) has been explored in previous works, primarily in NLP. Gururangan et al.~\cite{continued-pretraining} show improvements on domain-specific tasks (medical text data, legal documents, etc.) by continuing the pretraining task on RoBERTa~\cite{DBLP:journals/corr/abs-1907-11692} with in-domain data. Subsequent work~\cite{embert} investigated CoPT in multilingual BERT models to improve finetuned performance on languages that were unrepresented or under-represented in the pretraining data set. 

In this work, we investigate CoPT in the context of ASR with experiments on XLSR-53~\cite{xlsr-53}. XLSR-53 is a multilingual wav2vec 2.0 \cite{wav2vec} model trained on 56k hours of audio from 53 different languages (approximately 80\% of which is English data) which has demonstrated impressive performance when finetuned on various languages. We find that CoPT consistently improves performance across the languages we investigated. Since CoPT uses unlabeled data, we compare the performance of CoPT with semi-supervised training (SST)~\cite{DBLP:conf/icassp/MaMKS06,DBLP:conf/asru/VeselyHB13,DBLP:conf/icassp/ThomasSCH13,DBLP:conf/icassp/ManoharHPK18,DBLP:journals/speech/YuGWW10,wav2vec-sst}. SST uses a bootstrap ASR model to generate pseudo labels for the unlabeled data; this pseudo labeled data is filtered to discard lower quality data and combined with the labeled data, and the model retrained~\cite{DBLP:conf/interspeech/ZavaliagkosSCB98}.  

Our results show that CoPT performs as well, if not better than, SST in low-resource languages while being computationally cheaper to train and simpler to implement. In addition, we show that both CoPT and SST can be used together to further improve performance. While CoPT has been explored in NLP, this work is the first to demonstrate its effectiveness in low-resource ASR and compare it with SST, the standard approach to using unlabeled data in ASR acoustic modeling.

The organization of the paper is as follows: Section~\ref{expt-setup} details the data sets used and the training setup and process. Section~\ref{results} describes the various experiments and results, with commentary. We conclude with closing remarks and suggestions for best practices with CoPT in Section~\ref{discuss}.

\section{Experiment Setup}\label{expt-setup}
\subsection{Data}
This work is primarily focused on improving WERs on low-resource languages. In particular, we focus on Georgian, Farsi, Somali, and Tagalog ASR with speech data from IARPA's MATERIAL program~\cite{material17}; this program's goal is to develop methods to find, retrieve, and summarize speech and text content in response to English queries using minimal human-created resources. For the ASR component of the program, a small training set, the \textbf{build} set, typically consisting of approximately 40-80 hours of telephone conversational speech  is provided, as well as a mixed speech (conversational speech, broadcast news, and conversation) test set, the \textbf{analysis} set. Of the languages considered here, Tagalog has about 150hrs of data; the other three languages have between 36-65 hours of audio data. All results on MATERIAL languages are reported on their respective \textbf{analysis} set. In addition to these data sets we also use Georgian\footnote{\url{https://catalog.ldc.upenn.edu/LDC2016S12}} and Tagalog\footnote{\url{https://catalog.ldc.upenn.edu/LDC2016S13}} BABEL data sets, available from LDC. These data sets include a train as well as dev test set. BABEL language results are reported on their corresponding dev test set. Prior to continued pretraining or finetuning, all audio undergoes 3-day speed perturbation~\cite{KoPPK15}.

As unsupervised audio data, we download audio from YouTube,  primarily from YouTube channels of news broadcast stations in the language of interest. For each language, we gather exactly 1000 hours of Youtube for use in SST and CoPT. The preselection of our pseudo-labelling process results in using a subset of this data, as the process attempts to remove generated transcripts that are likely to be too noisy for training. Since in practice, there is no concern of noisy transcripts for CoPT due to its self-supervised pretraining task, we utilize the entire 1000 hour set of Youtube data for CoPT.

For language modeling, we augment the training transcripts with web crawled data in each target language. This crawled data is minimally "cleaned" by removing punctuation, digits, tokens outside of the target language, and lower casing if applicable. The crawled data is also used to expand the lexicon used by the ASR systems in each language by adding all words that exceed a minimum threshold to the lexicon.

\subsection{Training} \label{training}
Our training setup for XLSR-53 differs from the original training approach in \cite{xlsr-53} and is most similar to Vyas et al. \cite{vyas}. In particular, we use~\cite{kaldi} for data preparation and decoding, and the Espresso toolkit~\cite{espresso} to train the XLSR-53 with the LF-MMI \cite{lf-mmi} objective rather than the CTC \cite{CTC} objective. The only rationale for using this setup rather than the original wav2vec 2.0 training paradigm is that we used this setup with success in our MATERIAL program efforts. The pretraining setup closely follows the original wav2vec 2.0 pretraining approach. To accommodate our compute environment, we train only on utterances $\leq$ 10 seconds and use an effective batch size of 4 minutes of audio, rather than the 32 used in the wav2vec 2.0 large configuration. For all languages, the finetuning process uses grapheme modeling, and the XLSR-53 model is finetuned for 25,000 updates. We found that 50,000 updates provides little improvement for the doubling the training time. For CoPT, the model is trained for 100,000 updates. All parameters are fixed across all languages under consideration; we did not attempt to optimize parameter values on individual languages to improve performance. Following \cite{wav2vec}, for each configuration, we train 3 different seeds (though the 3 seeds remain constant across all runs) and select the model with the best results. 

\subsection{Semi-Supervised Training}
The SST training data set for each language is created by decoding the corresponding unlabeled YouTube audio data with the baseline finetuned XLSR-53 model, using a language model based on both the training and web crawl data. We discard any decoded utterances where more than 10\% of the transcript contains \texttt{<UNK>} tokens, and per-frame-log-likelihood is below the median log-likelihood per frame for all utterances. For the Youtube data used, this process yields 450-650 hours of additional pseudo-labelled audio for supervised training. 

\section{Experiments and results}\label{results}
For each language, we consider four different CoPT setups:
\begin{itemize}
\item None: The baseline setup, no CoPT
\item BUILD: Using the {\bf build} data for CoPT 
\item YouTube: Using the 1000 hour unlabeled in-language YouTube audio for CoPT
\item BUILD + YouTube: Using both the {\bf build} data and corresponding 1000 hour unlabeled in-language YouTube audio for CoPT
\end{itemize}
 In all four setups we start from the publicly available XLSR-53 pretrained model\footnote{\url{https://dl.fbaipublicfiles.com/fairseq/wav2vec/xlsr_53_56k.pt}}.
 
To compare performance with SST, for each of these four CoPT setups we also finetune with the {\bf build} and YouTube SST data. For easy comparison across experiments, we generate one set of semi-supervised training transcripts in each language for all the experiments using the baseline model. A more realistic approach in high performance and production deployments is to regenerate SST datasets after each SST pass until computational considerations exceed performance gains. To demonstrate performance from such a pipeline for CoPT, we take the best CoPT based model after finetuning with the {\bf build} data, generate another set of SST transcripts, and repeat the finetuning with the newly generated SST data set. These results are also included in Table~\ref{copt-results} (labeled SST Best).

\subsection{MATERIAL Results} \label{material-results}
The results for the CoPT experiments on the MATERIAL languages are detailed in Table~\ref{copt-results}. Across all languages, CoPT improves the performance of the XLSR-53 model,  with larger pretraining data sets containing the 1000 hours of YouTube data providing more improvement than using the {\bf build} data alone in CoPT. In particular, CoPT appears to prime the XLSR-53 model to be more receptive to finetuning due to better modeling of underlying target language phonology.
\begin{table*}[ht]
  \caption{WER results comparing CoPT with different pretraining data with and without SST}
  \label{copt-results}
  \centering
  \begin{tabular}{lllcccc}
    \toprule
    \textbf{CoPT}  & \textbf{Finetuning} & \textbf{Georgian} & \textbf{Farsi} & \textbf{Somali} & \textbf{Tagalog} \\
    \midrule
    None	& BUILD	                      & 18.7	& 30.7		& 51.1		& 34.6 \\ 
    None	& BUILD + SST	          & 18.0	& 27.9	& 	50.4  &  26.6 \\ \midrule
    BUILD	& BUILD	                      & 18.5    & 32.1  &  50.4  &  33.1 \\
    BUILD	& BUILD + SST	          & 17.7	& 27.5	& 50.2	& 26.1 \\
    YouTube	& BUILD	                      & 17.5	& 27.6	& 49.4	& 26.6 \\
    YouTube	& BUILD + SST	          & 17.7	& 27.3	& 49.9	& 25.8 \\
    BUILD + YouTube	& BUILD	              & \textbf{17.4}	& 27.2	& \textbf{48.9}	& 26.5 \\
    BUILD + YouTube	& BUILD + SST	  & 17.6	& 27.2	& 49.9	& 25.8 \\ \midrule
    BUILD + YouTube & BUILD + SST Best & 17.6 & \textbf{26.5} & 49.0 & \textbf{24.4} \\
    \bottomrule
  \end{tabular}
\end{table*}

Two results in  Table~\ref{copt-results} are of particular interest, CoPT None with BUILD + SST finetuning (row 2) and CoPT BUILD + YouTube with BUILD finetuning (row 7). The first is a standard SST pipeline implemented on the XLSR-53 model and the second is a straightforward implementation of CoPT on XLSR-53. Across all languages, we observe that CoPT matches or slightly outperforms SST. CoPT only requires continuing the pretraining task, which is much simpler than setting up a separate pseudo-labeling workflow. SST, on the other hand, requires decoding of 1000+ hours of audio followed by finetuning over the SST transcripts at a significant computational cost. Essentially, CoPT in pretrained models can replace SST while retaining the same performance with lower computational cost and effort.

The results with {\bf build} pretraining are mixed. In general, {\bf build} pretraining, comparing rows 1 and 3 in Table~\ref{copt-results},  improves performance.  However, Farsi shows a degradation in WER with  {\bf build} pretraining. It is not clear why we see the degradation, we do note that of the languages considered, Farsi has the smallest {\bf build} data set (~36 hours). CoPT with very small datasets may require a different set of meta parameters for optimal performance; as noted earlier the Section~\ref{training}, for consistency, we did not optimize parameters on individual languages. All CoPT and finetuning workflows and parameters are identical for all languages.

As mentioned previously, the baseline system was used to generate transcripts for SST. However, given that the best CoPT model without SST has better performance, we can regenerate transcripts, i.e. pseudo labels, for SST with the best CoPT model. The last row in Table~\ref{copt-results}, labeled BUILD + SST BEST, shows the results of finetuning the CoPT model with SST using more accurate transcripts. As shown, this gives further improvements on Farsi and Tagalog, with Georgian and Somali essentially the same as prior best results.

Finally, Table~\ref{genre-results} breaks down the MATERIAL results by genre, as mentioned earlier, the \textbf{analysis} test set consists of three different genres: telephone (conversational) speech (CS), news broadcasts (NB), and topical broadcasts (TB). The largest improvements are observed in the TB and NB genres in all languages. This is likely due to the fact that YouTube audio more closely matches TB and NB speech. We note that CS performance improves as well, though not to the same degree, even though the YouTube audio is quite different from conversational speech.
\begin{table}[ht]
  \caption{MATERIAL \textbf{analysis} WER by genre with finetuning using \textbf{build} only.}
  \label{genre-results}
  \centering
  \begin{tabular}{llccc}
    \toprule
    \textbf{Language}      & \textbf{CoPT}    & \textbf{CS} & \textbf{NB} & \textbf{TB}            \\
    \midrule
    \multirow{ 2}{*}{Georgian} & None         & 33.8 & 11.9 & 19.0   \\
                               & BUILD + YT   & 32.7 & 10.7 & 17.4   \\
    \midrule
    \multirow{ 2}{*}{Farsi}  & None         & 41.5 & 26.6 & 30.8   \\
                             & BUILD + YT   & 38.2 & 23.1 & 27.2   \\
    \midrule
    \multirow{ 2}{*}{Somali}  & None         & 59.0 & 47.1 & 52.1   \\
                             & BUILD + YT   & 56.4 & 44.6 & 49.5   \\
    \midrule
    \multirow{ 2}{*}{Tagalog}  & None         & 45.2 & 25.3 & 35.6   \\
                              & BUILD + YT   & 43.5 & 19.0 & 27.6   \\
    \bottomrule
  \end{tabular}
\end{table}

\subsection{BABEL Results}
From the results in Section~\ref{material-results}, it is clear that in-language and in-domain CoPT help in low resource ASR. To further understand the impact of out-of-domain CoPT, we start from the language-specific YouTube CoPT models and finetune on the Georgian and Tagalog BABEL data sets and compare performance to finetuned models without CoPT. The language-specific YouTube CoPT models used here are the same models that are finetuned to generate rows 5 and 6 in Table~\ref{copt-results} for Georgian and Tagalog. Note that BABEL data sets include conversational speech data only, whereas YouTube data more closely matches broadcast news than conversational, representing in-language but out of domain data. The results of these experiments are detailed in Table~\ref{babel-results} and mirror the behavior observed in Table~\ref{genre-results} on the MATERIAL \textbf{analysis} CS subsets, with diminished gains when using out-of-domain CoPT.
\begin{table}[ht]
  \caption{BABEL dev set WER with CoPT with finetuning using corresponding train set.}
  \label{babel-results}
  \centering
  \begin{tabular}{llc}
    \toprule
    \textbf{Language}      & \textbf{CoPT}  & \textbf{WER}            \\ \midrule
    \multirow{ 2}{*}{Georgian} & None         & 31.9   \\
                               & YT   & 30.7   \\
    \midrule
    \multirow{ 2}{*}{Tagalog} & None         & 36.3  \\
                              & YT   & 35.2   \\
    \bottomrule
  \end{tabular}
\end{table}

\section{Discussion}\label{discuss}

In this paper, we present results demonstrating that CoPT provides similar to, or better, results than SST on large pretrained Transformer models, in particular, XLSR-53. CoPT is computationally efficient and simpler to implement compared to SST. We further show that CoPT and SST are complementary and that using that both in a pipeline, where CoPT models generate transcripts for SST, results in additional improvements in WER. CoPT performs best with in-domain data but this is not a hard requirement; out-of-domain data also results in performance improvements albeit smaller than with in-domain data. Based on these results, our recommendation for XLSR-53 like models is to apply CoPT first and then, compute and time permitting, implement SST using the CoPT based model to generate SST transcripts.

\section{Acknowledgements}

This research is based upon work supported by the Office of the Director of National Intelligence (ODNI), Intelligence Advanced Research Projects Activity (IARPA), via AFRL Contract \#FA8650-17-C-9116. The views and conclusions contained herein are those of the authors and should not be interpreted as necessarily representing the official policies or endorsements, either expressed or implied, of the ODNI, IARPA, or the U.S. Government. The U.S. Government is authorized to reproduce and distribute reprints for Governmental purposes notwithstanding any copyright annotation thereon.

\bibliographystyle{IEEEtran}
\bibliography{mybib}

\begin{thebibliography}{10}
\providecommand{\url}[1]{#1}
\csname url@samestyle\endcsname
\providecommand{\newblock}{\relax}
\providecommand{\bibinfo}[2]{#2}
\providecommand{\BIBentrySTDinterwordspacing}{\spaceskip=0pt\relax}
\providecommand{\BIBentryALTinterwordstretchfactor}{4}
\providecommand{\BIBentryALTinterwordspacing}{\spaceskip=\fontdimen2\font plus
\BIBentryALTinterwordstretchfactor\fontdimen3\font minus
  \fontdimen4\font\relax}
\providecommand{\BIBforeignlanguage}[2]{{%
\expandafter\ifx\csname l@#1\endcsname\relax
\typeout{** WARNING: IEEEtran.bst: No hyphenation pattern has been}%
\typeout{** loaded for the language `#1'. Using the pattern for}%
\typeout{** the default language instead.}%
\else
\language=\csname l@#1\endcsname
\fi
#2}}
\providecommand{\BIBdecl}{\relax}
\BIBdecl

\bibitem{radford2018improving}
A.~Radford, K.~Narasimhan, T.~Salimans, and I.~Sutskever, ``Improving language
  understanding by generative pre-training,'' 2018.

\bibitem{radford2019language}
A.~Radford, J.~Wu, R.~Child, D.~Luan, D.~Amodei, I.~Sutskever \emph{et~al.},
  ``Language models are unsupervised multitask learners,'' \emph{OpenAI blog},
  vol.~1, no.~8, p.~9, 2019.

\bibitem{DBLP:journals/corr/abs-1810-04805}
\BIBentryALTinterwordspacing
J.~Devlin, M.~Chang, K.~Lee, and K.~Toutanova, ``{BERT:} pre-training of deep
  bidirectional transformers for language understanding,'' \emph{CoRR}, vol.
  abs/1810.04805, 2018. [Online]. Available:
  \url{http://arxiv.org/abs/1810.04805}
\BIBentrySTDinterwordspacing

\bibitem{vit}
\BIBentryALTinterwordspacing
A.~Dosovitskiy, L.~Beyer, A.~Kolesnikov, D.~Weissenborn, X.~Zhai,
  T.~Unterthiner, M.~Dehghani, M.~Minderer, G.~Heigold, S.~Gelly, J.~Uszkoreit,
  and N.~Houlsby, ``An image is worth 16x16 words: Transformers for image
  recognition at scale,'' in \emph{9th International Conference on Learning
  Representations, {ICLR} 2021, Virtual Event, Austria, May 3-7, 2021}.\hskip
  1em plus 0.5em minus 0.4em\relax OpenReview.net, 2021. [Online]. Available:
  \url{https://openreview.net/forum?id=YicbFdNTTy}
\BIBentrySTDinterwordspacing

\bibitem{wav2vec}
A.~Baevski, Y.~Zhou, A.~Mohamed, and M.~Auli, ``wav2vec 2.0: A framework for
  self-supervised learning of speech representations,'' \emph{Advances in
  Neural Information Processing Systems}, vol.~33, pp. 12\,449--12\,460, 2020.

\bibitem{xlm}
G.~Lample and A.~Conneau, ``Cross-lingual language model pretraining,'' in
  \emph{NeurIPS}, 2019.

\bibitem{xlsr-53}
A.~Conneau, A.~Baevski, R.~Collobert, A.~Mohamed, and M.~Auli, ``{Unsupervised
  Cross-Lingual Representation Learning for Speech Recognition},'' in
  \emph{Proc. Interspeech 2021}, 2021, pp. 2426--2430.

\bibitem{wu-dredze}
\BIBentryALTinterwordspacing
S.~Wu and M.~Dredze, ``Are all languages created equal in multilingual
  {BERT}?'' in \emph{Proceedings of the 5th Workshop on Representation Learning
  for NLP}.\hskip 1em plus 0.5em minus 0.4em\relax Online: Association for
  Computational Linguistics, Jul. 2020, pp. 120--130. [Online]. Available:
  \url{https://aclanthology.org/2020.repl4nlp-1.16}
\BIBentrySTDinterwordspacing

\bibitem{biobert}
J.~Lee, W.~Yoon, S.~Kim, D.~Kim, S.~Kim, C.~H. So, and J.~Kang, ``{BioBERT: a
  pre-trained biomedical language representation model for biomedical text
  mining},'' \emph{Bioinformatics}, vol.~36, no.~4, pp. 1234--1240, 09 2019.

\bibitem{bertweet}
\BIBentryALTinterwordspacing
D.~Q. Nguyen, T.~Vu, and A.~Tuan~Nguyen, ``{BERT}weet: A pre-trained language
  model for {E}nglish tweets,'' in \emph{Proceedings of the 2020 Conference on
  Empirical Methods in Natural Language Processing: System
  Demonstrations}.\hskip 1em plus 0.5em minus 0.4em\relax Online: Association
  for Computational Linguistics, Oct. 2020, pp. 9--14. [Online]. Available:
  \url{https://aclanthology.org/2020.emnlp-demos.2}
\BIBentrySTDinterwordspacing

\bibitem{scibert}
\BIBentryALTinterwordspacing
I.~Beltagy, K.~Lo, and A.~Cohan, ``{S}ci{BERT}: A pretrained language model for
  scientific text,'' in \emph{Proceedings of the 2019 Conference on Empirical
  Methods in Natural Language Processing and the 9th International Joint
  Conference on Natural Language Processing (EMNLP-IJCNLP)}.\hskip 1em plus
  0.5em minus 0.4em\relax Hong Kong, China: Association for Computational
  Linguistics, Nov. 2019, pp. 3615--3620. [Online]. Available:
  \url{https://aclanthology.org/D19-1371}
\BIBentrySTDinterwordspacing

\bibitem{continued-pretraining}
\BIBentryALTinterwordspacing
S.~Gururangan, A.~Marasovi{\'c}, S.~Swayamdipta, K.~Lo, I.~Beltagy, D.~Downey,
  and N.~A. Smith, ``Don{'}t stop pretraining: Adapt language models to domains
  and tasks,'' in \emph{Proceedings of the 58th Annual Meeting of the
  Association for Computational Linguistics}.\hskip 1em plus 0.5em minus
  0.4em\relax Online: Association for Computational Linguistics, Jul. 2020, pp.
  8342--8360. [Online]. Available:
  \url{https://aclanthology.org/2020.acl-main.740}
\BIBentrySTDinterwordspacing

\bibitem{DBLP:journals/corr/abs-1907-11692}
\BIBentryALTinterwordspacing
Y.~Liu, M.~Ott, N.~Goyal, J.~Du, M.~Joshi, D.~Chen, O.~Levy, M.~Lewis,
  L.~Zettlemoyer, and V.~Stoyanov, ``Roberta: {A} robustly optimized {BERT}
  pretraining approach,'' \emph{CoRR}, vol. abs/1907.11692, 2019. [Online].
  Available: \url{http://arxiv.org/abs/1907.11692}
\BIBentrySTDinterwordspacing

\bibitem{embert}
\BIBentryALTinterwordspacing
Z.~Wang, K.~K, S.~Mayhew, and D.~Roth, ``Extending multilingual {BERT} to
  low-resource languages,'' in \emph{Findings of the Association for
  Computational Linguistics: EMNLP 2020}.\hskip 1em plus 0.5em minus
  0.4em\relax Online: Association for Computational Linguistics, Nov. 2020, pp.
  2649--2656. [Online]. Available:
  \url{https://aclanthology.org/2020.findings-emnlp.240}
\BIBentrySTDinterwordspacing

\bibitem{DBLP:conf/icassp/MaMKS06}
J.~Z. Ma, S.~Matsoukas, O.~Kimball, and R.~M. Schwartz, ``Unsupervised training
  on large amounts of broadcast news data,'' in \emph{{ICASSP} 2006, Toulouse,
  France, May 14-19}.\hskip 1em plus 0.5em minus 0.4em\relax {IEEE}, 2006, pp.
  1056--1059.

\bibitem{DBLP:conf/asru/VeselyHB13}
K.~Vesel{\'{y}}, M.~Hannemann, and L.~Burget, ``Semi-supervised training of
  deep neural networks,'' in \emph{2013 {IEEE} Workshop on Automatic Speech
  Recognition and Understanding, Olomouc, Czech Republic, December 8-12}.\hskip
  1em plus 0.5em minus 0.4em\relax {IEEE}, 2013, pp. 267--272.

\bibitem{DBLP:conf/icassp/ThomasSCH13}
S.~Thomas, M.~L. Seltzer, K.~Church, and H.~Hermansky, ``Deep neural network
  features and semi-supervised training for low resource speech recognition,''
  in \emph{{ICASSP} 2013, Vancouver, BC, Canada, May 26-31}.\hskip 1em plus
  0.5em minus 0.4em\relax {IEEE}, 2013, pp. 6704--6708.

\bibitem{DBLP:conf/icassp/ManoharHPK18}
V.~Manohar, H.~Hadian, D.~Povey, and S.~Khudanpur, ``Semi-supervised training
  of acoustic models using lattice-free {MMI},'' in \emph{{ICASSP} 2018,
  Calgary, AB, Canada, April 15-20}.\hskip 1em plus 0.5em minus 0.4em\relax
  {IEEE}, 2018, pp. 4844--4848.

\bibitem{DBLP:journals/speech/YuGWW10}
K.~Yu, M.~J.~F. Gales, L.~Wang, and P.~C. Woodland, ``Unsupervised training and
  directed manual transcription for {LVCSR},'' \emph{Speech Communication},
  vol.~52, no. 7-8, pp. 652--663, 2010.

\bibitem{wav2vec-sst}
Q.~Xu, A.~Baevski, T.~Likhomanenko, P.~Tomasello, A.~Conneau, R.~Collobert,
  G.~Synnaeve, and M.~Auli, ``Self-training and pre-training are complementary
  for speech recognition,'' in \emph{{IEEE} International Conference on
  Acoustics, Speech and Signal Processing, {ICASSP} 2021, Toronto, ON, Canada,
  June 6-11, 2021}.\hskip 1em plus 0.5em minus 0.4em\relax {IEEE}, 2021, pp.
  3030--3034.

\bibitem{DBLP:conf/interspeech/ZavaliagkosSCB98}
\BIBentryALTinterwordspacing
G.~Zavaliagkos, M.~Siu, T.~Colthurst, and J.~Billa, ``Using untranscribed
  training data to improve performance,'' in \emph{The 5th International
  Conference on Spoken Language Processing, Sydney Convention Centre, Sydney,
  Australia, 30th November - 4th December 1998}.\hskip 1em plus 0.5em minus
  0.4em\relax {ISCA}, 1998. [Online]. Available:
  \url{http://www.isca-speech.org/archive/icslp_1998/i98_1007.html}
\BIBentrySTDinterwordspacing

\bibitem{material17}
``Machine {T}ranslation for {E}nglish {R}etrieval of {I}nformation in {A}ny
  {L}anguage ({MATERIAL}) program,''
  \url{https://www.iarpa.gov/index.php/research-programs/material}.

\bibitem{KoPPK15}
\BIBentryALTinterwordspacing
T.~Ko, V.~Peddinti, D.~Povey, and S.~Khudanpur, ``Audio augmentation for speech
  recognition,'' in \emph{{INTERSPEECH} 2015, Dresden, Germany, September
  6-10}, 2015, pp. 3586--3589. [Online]. Available:
  \url{http://www.isca-speech.org/archive/interspeech_2015/i15_3586.html}
\BIBentrySTDinterwordspacing

\bibitem{vyas}
A.~Vyas, S.~Madikeri, and H.~Bourlard, ``{Comparing CTC and LFMMI for
  Out-of-Domain Adaptation of wav2vec 2.0 Acoustic Model},'' in \emph{Proc.
  Interspeech 2021}, 2021, pp. 2861--2865.

\bibitem{kaldi}
D.~Povey, A.~Ghoshal, G.~Boulianne, L.~Burget, O.~Glembek, N.~Goel,
  M.~Hannemann, P.~Motlicek, Y.~Qian, P.~Schwarz, J.~Silovsky, G.~Stemmer, and
  K.~Vesely, ``The {K}aldi {S}peech {R}ecognition {T}oolkit,'' in \emph{IEEE
  2011 Workshop on Automatic Speech Recognition and Understanding}.\hskip 1em
  plus 0.5em minus 0.4em\relax IEEE Signal Processing Society, Dec. 2011, iEEE
  Catalog No.: CFP11SRW-USB.

\bibitem{espresso}
Y.~Wang, T.~Chen, H.~Xu, S.~Ding, H.~Lv, Y.~Shao, N.~Peng, L.~Xie, S.~Watanabe,
  and S.~Khudanpur, ``Espresso: A fast end-to-end neural speech recognition
  toolkit,'' in \emph{2019 IEEE Automatic Speech Recognition and Understanding
  Workshop (ASRU)}, 2019.

\bibitem{lf-mmi}
D.~Povey, V.~Peddinti, D.~Galvez, P.~Ghahremani, V.~Manohar, X.~Na, Y.~Wang,
  and S.~Khudanpur, ``Purely sequence-trained neural networks for {ASR} based
  on lattice-free {MMI},'' in \emph{Interspeech 2016, 17th Annual Conference of
  the International Speech Communication Association, San Francisco, CA, USA,
  September 8-12, 2016}, N.~Morgan, Ed.\hskip 1em plus 0.5em minus 0.4em\relax
  {ISCA}, 2016, pp. 2751--2755.

\bibitem{CTC}
A.~Graves, S.~Fern\'{a}ndez, F.~Gomez, and J.~Schmidhuber, ``Connectionist
  temporal classification: Labelling unsegmented sequence data with recurrent
  neural networks,'' in \emph{Proceedings of the 23rd International Conference
  on Machine Learning}, ser. ICML '06.\hskip 1em plus 0.5em minus 0.4em\relax
  New York, NY, USA: Association for Computing Machinery, 2006, p. 369–376.

\end{thebibliography}
\end{document}